\documentclass[letterpaper]{article} 
\usepackage{aaai2026}  
\usepackage{times}  
\usepackage{helvet}  
\usepackage{courier}  
\usepackage[hyphens]{url}  
\usepackage{graphicx} 
\def\UrlFont{\rm}  
\usepackage{natbib}  
\usepackage{caption} 
\usepackage{xcolor} 
\usepackage{algorithm}
\usepackage{algorithmic}
\usepackage{amsmath}
\usepackage{newfloat}
\usepackage{listings}
\usepackage{threeparttable} 
\usepackage{subcaption}  
\DeclareCaptionStyle{ruled}{labelfont=normalfont,labelsep=colon,strut=off} 
\floatstyle{ruled}
\newfloat{listing}{tb}{lst}{}
\floatname{listing}{Listing}
\title{Realistic Synthetic Household Data Generation at Scale}
\author {
    Siddharth Singh \textsuperscript{\rm 1}\equalcontrib,
    Ifrah Idrees \textsuperscript{\rm 1} \equalcontrib,
    Abraham Dauhajre \textsuperscript{\rm 1}
}
\affiliations {
    \textsuperscript{\rm 1}Amazon Lab 126\\
    hartsid@amazon.com, idreesio@amazon.com}

\begin{document}

\maketitle

    
    

\begin{abstract}
Recent advancements in foundation models have catalyzed research in Embodied AI to develop interactive agents capable of environmental reasoning and interaction. Developing such agents requires diverse, large-scale datasets. Prior frameworks generate synthetic data for long-term human-robot interactions and 3D environments but fail to model the bidirectional influence between human behavior and household environments. Our proposed generative framework creates household datasets at scale through loosely coupled generation of long-term human-robot interactions and environments. Human personas influence environment generation, while environment schematics and semantics shape human-robot interactions.

The generated 3D data includes rich static context such as object and environment semantics, as well as temporal context capturing human and agent behaviors over extended periods. 
Our flexible tool allows users to define dataset characteristics via natural language prompts, enabling configuration of both environment and human activity data through natural language specifications. The tool can create variations of user-defined configurations, thus enabling scalable data generation.
 
We validate our framework through comprehensive statistical evaluation using multi-modal embeddings and three key metrics: cosine similarity analysis, mutual information gain, intervention analysis, and iterative improvement validation. Statistical comparisons demonstrates good alignment with real-world datasets (HOMER) showing high cosine similarity values (0.60), while comparisons with synthetic datasets (Wang et al.) show moderate alignment (0.27). Intervention analysis across age, organization, and sleep pattern modifications shows statistically significant effects (p $<$ 0.001) with large effect sizes (Cohen's d = 0.51-1.12), confirming that bidirectional coupling successfully translates persona characteristics into measurable differences in both environmental configurations and behavioral patterns.All these contributions will enable the development and testing of household smart devices at scale.
\end{abstract}

\section{Introduction}

Training robots and devices for household environments presents a fundamental technical challenge: modeling and understanding the bidirectional relationship between human behavioral patterns and environmental configurations. This challenge is particularly critical for robots, which must operate safely and effectively in diverse household settings. The challenge necessitates addressing three core technical problems: (i) \textbf{spatio-temporal dependencies} - capturing how human activities influence object placement, room organization, and environmental semantics over extended time periods, (ii) \textbf{bidirectional coupling} - modeling the mutual influence between human behaviors and environmental configurations, and (iii) \textbf{semantic relationships} - ensuring coherent connections between persona characteristics, environmental affordances, and behavioral patterns.

Current robots struggle with these dynamics, as they require training data that simultaneously represents: (i) static environmental properties (object affordances, spatial relationships, semantic labels), and (ii) temporal patterns of human-environment interactions (daily routines, object manipulation sequences, long-term space utilization). Without this understanding, robots cannot effectively anticipate human needs, adapt to dynamic household configurations, or provide meaningful assistance. The robotics industry faces particular challenges in this domain, as household robots—including cleaning robots, assistive robots, and smart home systems—must operate reliably across diverse family dynamics and home configurations while ensuring user safety and satisfaction.

Most current approaches attempt to solve the data generation problem for training robotics using decoupled methodologies. This is particularly problematic for robots, which require comprehensive understanding of human-environment interactions to function effectively in household settings. Previous works for generating environment schematics for indoor environments have employed algorithmic and generative approaches. Additionally, generative methods that work in conjunction with large language models can process unstructured data about human personas and behaviors but lack granular activity-level modeling. Similarly, existing methodologies generate human activity data while considering environmental context but fail to: (i) inform environment generation processes, and (ii) account for environmental details at the granular level of regions, furniture, and asset placement.

Our framework addresses these limitations by performing bidirectional generation of both environmental configurations and human behavioral activities through a loosely coupled architecture that enables mutual information exchange. Building upon our previous work on realistic simulation of daily human activity \cite{10309457}, the framework employs large language models with iterative refinement mechanisms to ensure semantic consistency between generated environments and behaviors. Specifically, the presence of environmental assets (e.g., dishwasher, gaming console) and regions (home office, art studio, etc.) guides activity generation modules to synthesize corresponding behaviors (e.g., dishwasher loading, video gaming, coding, pottery, etc.). Conversely, generated activities (e.g., laundry tasks, sketching) inform the environment generation module to include semantically relevant assets (e.g., laundry baskets, art supplies). Through this bidirectional coupling, our framework ensures that generated activity data is both semantically rich and environmentally grounded.

This research directly addresses the data scarcity challenge facing the robotics industry, where companies developing household robots require diverse, large-scale datasets to ensure their products can operate safely and effectively across varied household environments and family dynamics. This research provides the following key contributions:
\begin{figure*}[h]
    \centering
    \includegraphics[width=0.9\textwidth]{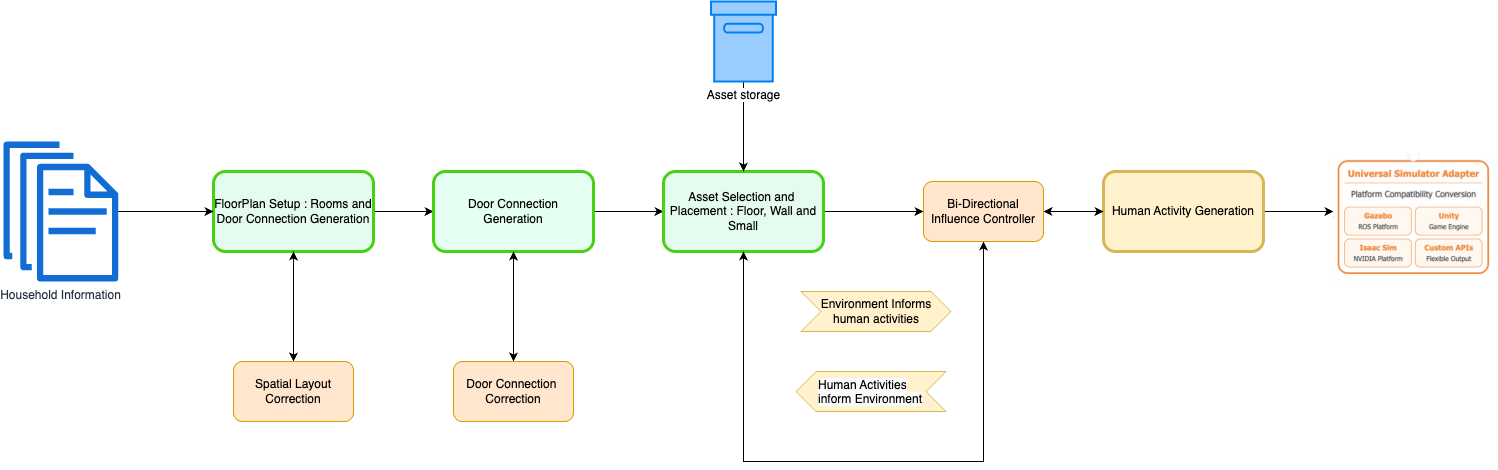}
    \caption{Framework Pipeline Overview: Our bidirectional generation framework comprises three primary modules operating in an iterative refinement cycle. The Environment Schematic Generator produces 3D household layouts based on persona-driven requirements. The Human Activity and HRI Generator synthesizes temporally consistent behavior sequences. The Bidirectional Influence Controller orchestrates iterative information exchange between the modules.}
    \label{fig:pipeline}
\end{figure*}

\begin{enumerate}
\item A novel framework for human activity and  human-robot interaction synthetic data generation with temporal consistency
\item A persona-driven framework for synthetic indoor environment generation based on environment schematics and household member characteristics
\item Bidirectional coupling mechanisms between generation frameworks to produce rich and grounded synthetic household data
\item Comprehensive statistical validation demonstrating how bidirectional influence enhances household data generation quality through multi-modal embedding analysis
\item Sim-to-real validation experiments confirming that generated datasets align with real-world data distributions
\end{enumerate}

\begin{figure*}[t]
    \centering
    \includegraphics[width=0.9\textwidth]{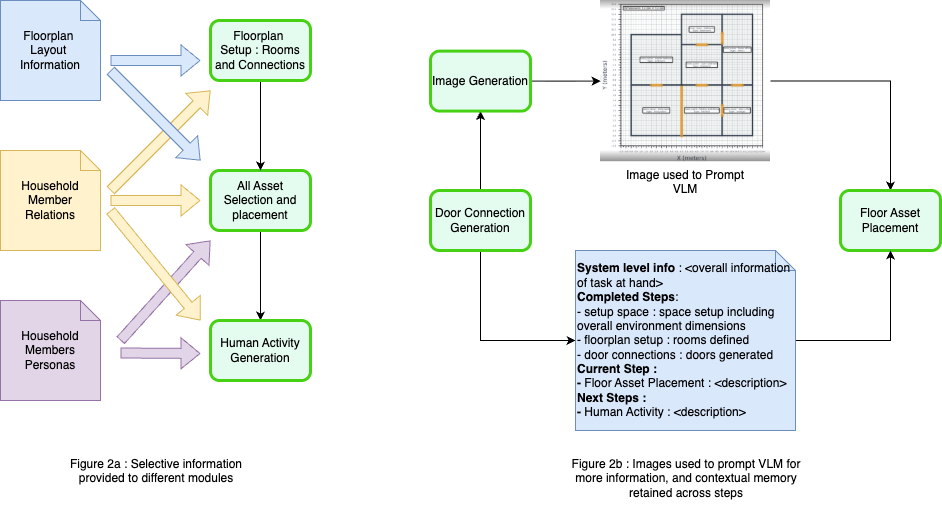}
    \caption{Input Specification and Contextual Memory Framework: Our system accepts structured natural language descriptions of household member personas and environmental constraints. The framework maintains contextual memory across the pipeline, providing the LLM with context regarding task requirements and completed steps to reduce hallucination issues.}
    \label{fig:input_specification}
\end{figure*}
\section{Related Work}



\textbf{Synthetic Data Generation for Indoor Environments:} Historical approaches to indoor environment generation have predominantly employed algorithmic and procedural techniques. Early methodologies such as ProcGen utilized constrained asset databases, primarily sourcing from repositories like Objaverse, to generate structured indoor layouts through rule-based systems \cite{Dai31122023}. These approaches, while computationally efficient, suffered from limited semantic diversity and restricted configurability, resulting in environments that lacked the nuanced characteristics necessary for realistic household modeling.
The emergence of Large Language Models (LLMs) and Vision-Language Models (VLMs) has enabled open-vocabulary generation techniques that transcend the limitations of constrained datasets. Contemporary approaches leverage the generative capabilities of foundation models to synthesize both environmental descriptions 
from natural language specifications. These methodologies demonstrate enhanced semantic richness and configurability compared to their algorithmic predecessors. Recent frameworks such as Holodeck employ large language models to generate environment schematics based on high-level descriptions of environmental characteristics and household member profiles \cite{yang2024holodeck}. These approaches demonstrate significant improvements in semantic coherence and user controllability, enabling the generation of contextually appropriate indoor environments through natural language interfaces. 

\textbf{Human Behavior Modeling and Generation:} Recent advances in scene understanding and human behavior modeling have produced several significant contributions. Behavior-1K introduces a comprehensive dataset encompassing 1,000 human behavior categories with a multi-modal learning framework for behavior recognition and generation \cite{pmlr-v205-li23a}. The framework integrates textual and visual modalities to model human actions but lacks environmental adaptation capabilities and bidirectional influence mechanisms.

Traditional human activity synthesis has relied on probabilistic modeling techniques that operate on predefined activity taxonomies. Early frameworks employed finite state machines and Markov models to generate activity sequences from constrained activity sets \cite{inproceedings}. While these methods provided temporal consistency, they lacked the flexibility to incorporate persona-specific behavioral patterns and environmental context, limiting their applicability to diverse household scenarios. Our previous work \cite{10309457} established foundational approaches for realistic simulation of daily human activity using constraint propagation algorithms, which inform the current framework's approach to activity generation.

Recent work in generating synthetic human activity generation for embodied AI has taken diverse approaches.The Dynamic Scene Generation research by \citeauthor{wang2024dynamic} uses a hierarchical framework with Large Language Models (LLMs) to generate human activity schedules and corresponding object relocations. It applies these to existing static scenes. While effective in modeling temporal scene dynamics, this approach does not incorporate personality-driven factors in scene generation or account for human behavioral influences on environmental configuration. 

Another framework by \citeauthor{Li} uses a hierarchical structure (Needs, Tasks, Activities) with a Behavior Planner to generate real-time, personality-driven behaviors in predefined 3D environments, producing behavior sequences for virtual humans in VR/AR settings. While X's Day \cite{Li} prioritizes dynamic, personality-aligned behaviors for interactive applications, our framework focuses on generating integrated environment-behavior datasets that model long-term household dynamics.

\textbf{Gap Analysis:} Existing methodologies for both Environment and Human Generation exhibit fundamental limitations in addressing the bidirectional relationship between human behavior and environmental adaptation. Current approaches treat environment generation and human activity synthesis as independent processes, failing to capture the complex interdependencies that characterize realistic household dynamics. Behavior-1K emphasizes behavior recognition without environmental interaction modeling, while Dynamic Scene Graph Generation handles temporal scene relationships but lacks integration with human behavioral factors.

The field requires frameworks that integrate comprehensive behavior understanding with dynamic scene adaptation, particularly systems capable of generating environments that respond to and influence human behavior patterns. This integration is essential for creating realistic synthetic datasets that accurately reflect the complex human-environment interactions characteristic of household settings.

\section{Methodology}

\subsection{Framework Overview}
Our framework addresses the fundamental challenge of generating realistic household data by establishing a bidirectional coupling between environment generation and human activity synthesis. Unlike previous approaches that treat these components independently, our system creates a feedback loop where human personas inform environmental characteristics, and environmental semantics guide activity generation.

\textbf{System Architecture:} The framework comprises four primary modules operating in an iterative refinement cycle: 
\begin{enumerate}
\item \textit{Environment Schematic Generator}: Produces 3D household layouts with semantic object selection and placement based on persona-driven requirements
\item \textit{Human Activity and HRI Generator}: Synthesizes temporally consistent behavior sequences grounded in environmental affordances  
\item \textit{Bidirectional Influence Controller}: Orchestrates the iterative exchange of information between the two generation modules to ensure semantic consistency
\item \textit{Universal Simulator Adapter}: Converts intermediate representations into formats compatible with various simulation environments, enabling simulator-agnostic deployment while maintaining semantic consistency
\end{enumerate}

Figure~1 illustrates all components of the pipeline at a high level. 
The iterative bidirectional influence process continues for a predetermined number of iterations until convergence criteria are met: (i) no new activity additions are expected based on environmental context, and (ii) no environmental modifications are anticipated due to generated activities. Detailed specifications of this iterative process and refinement criteria are provided in subsequent sections.
\begin{figure*}[!t]
    \centering
    \includegraphics[width=0.8\textwidth]{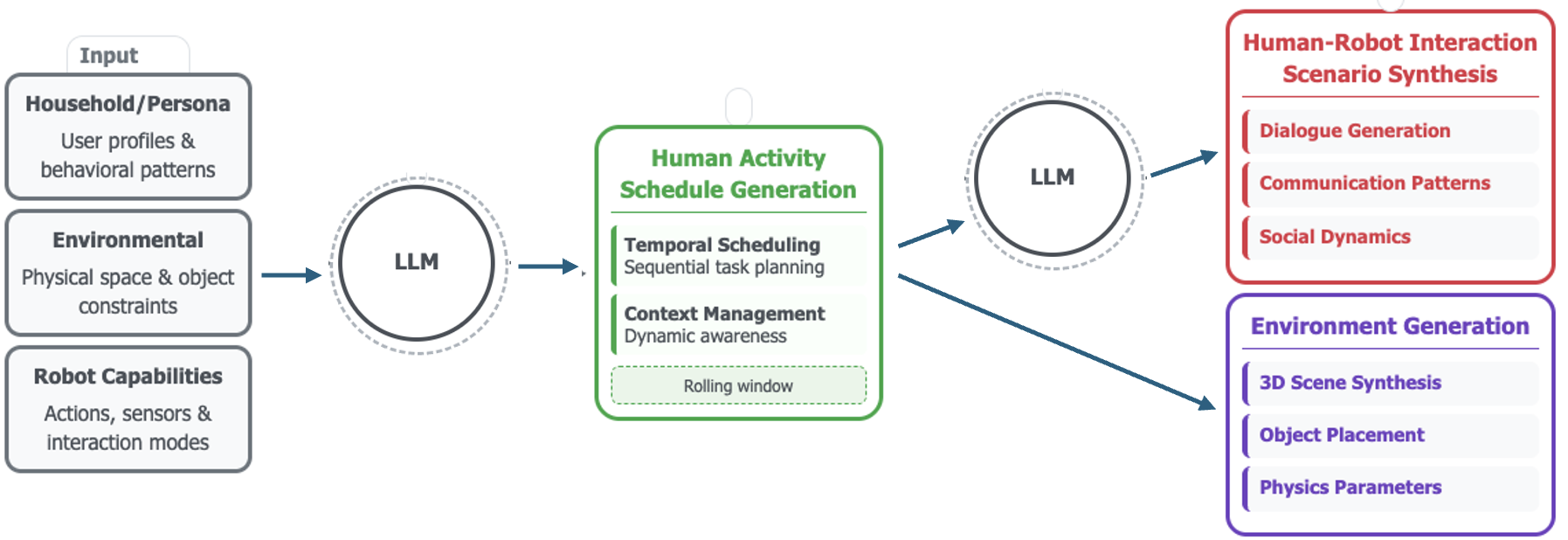}  
    \caption{Human Activity and Interaction Generation Pipeline}
    \label{fig:mainfig}
\end{figure*}

\textbf{Input Specification:} The system accepts natural language descriptions of household member personas, encompassing demographic information, lifestyle preferences, daily routines, and behavioral characteristics. Additionally, users can also specify high-level environmental constraints (house type, room configurations, functional requirements).

In contrast to the previous methods which provide free-form input text across all stages of generation, we find providing information in a structured manner helps the LLM generate better schematics and semantics. The input information is supplied to separate steps differently based on the task the step performs in the generation pipeline. Morevoer, we supplement visual information using intermediate outputs for subsequent steps. Steps such as door connection generation and object placement benefit from visual information about the floor-plan generated by previous steps.

Further, we maintain contextual memory across the entire pipeline, updated at each step. We observe that providing the LLM with context regarding: (i) the task being performed, (ii) pipeline steps, (iii) work completed by previous steps, and (iv) current step requirements, substantially reduces hallucination issues. Figure~2 provides information about input specification and maintenance of contextual memory usage.

\textbf{Scalability and Variation Generation:} The framework supports large-scale dataset creation through two primary mechanisms: (i) LLM parameter manipulation (temperature, top\_p, top\_k) for generation variability, and (ii) established algorithmic methods adapted from frameworks such as Holodeck for asset selection diversification and spatial optimization. This enables scalable dataset generation while preventing mode collapse and maintaining semantic coherence and sufficient diversity for embodied AI training.


\subsection{Human Activity and HRI Data Generation}

Building upon our previous work on realistic simulation of daily human activity \cite{10309457}, we developed a novel framework to generate contextually rich, temporally consistent human-activity and human-robot-interaction scenarios. Our framework extends the state-of-the-art approaches in generative agents to create realistic human activities and interaction spanning days, and weeks for specified personas, while enabling co-authoring capabilities. Our framework includes a Large Language Model (LLM) based implementation with a least-to-most prompt tuning approach. It further includes a rolling window context mechanism to maintain event consistency, ensuring logical progression of activities and interactions. Figure 3 describes this process in detail. It does so while maintaining coherent persona behaviors throughout. 


Our research introduces a robust framework for generating realistic human activities and robot interactions through a novel hierarchical decomposition approach. The framework leverages advanced language models and structured prompt engineering techniques to create coherent, contextually appropriate scenarios for human-robot interaction simulation. At the core of our approach lies a three-stage hierarchical decomposition strategy that breaks down the complex task of human activity generation into manageable, sequential components. Each subsequent task builds upon the outputs of previous stages, ensuring consistency and coherence in the generated scenarios.

The first stage focuses on activity generation, using household member profiles, environmental constraints, temporal parameters, and robot capabilities. This stage produces structured activity sequences that maintain both temporal and spatial consistency. By considering multiple contextual factors, our system generates realistic daily routines reflecting the nuances of different household dynamics.

The second stage synthesizes natural interactions and dialogues between humans and robots. This component processes the activity sequences from the first stage and enriches them with contextually appropriate conversations, accounting for social dynamics, cultural factors, and the specific characteristics of household members. The resulting dialogues maintain coherence while reflecting realistic human-robot communication patterns.

The third stage implements a universal simulator adapter, addressing one of the key challenges in human-robot interaction simulation: platform compatibility. This component converts our intermediate representation into formats compatible with various simulation environments, making our framework simulator-agnostic. This adaptation layer maintains semantic consistency while enabling flexible deployment across different platforms.

Our framework lets the researchers guide the generation of the synthetic data using natural language. The researchers can describe the scenario in an abstract manner while letting our framework fill in naturalistic details and assemble a structured synthetic dataset. For example, describing a human's persona — career, fitness preferences, hobbies, etc. — guides the kind of activities and interactions that the LLM will include in the dataset. This significantly expands the diversity and scale of testing scenarios available for development while reflecting the complexity of real-world human-robot interactions. 

\subsection{Environment Schematic Generation}

The environment generation module follows similar architectural patterns as the previous works while addressing several limitations:

\begin{enumerate}
\item \textbf{Asset Database Flexibility:} Most previous works have had a tight coupling with existing asset databases such as Objaverse. We find that swapping out asset databases provides tremendous benefits as long as correct metadata is available: (i) asset description, (ii) asset dimensions, (iii) asset pivot point, and (iv) asset image (optional but beneficial for enhanced generation quality).

\item \textbf{Room Layout Error Mitigation:} We address common room layout errors including nested rooms (rooms within rooms) and disconnected room configurations post-generation by algorithmic methods and correct them.

\item \textbf{Realistic Door Connection Generation:} Door connection generation is guided by LLM recommendations for based of the rooms being connected. For example, open concept layout floorplans often have entire walls removed between dining and living area. 
\end{enumerate}

Further, the integration of human persona and human activity information in the environment generation module helps guide the generation in several ways : 

\begin{enumerate}
\item \textbf{Personalized Room Assignment:} Each household member is assigned personal sleeping room.
\item \textbf{Occupation-Based Space Generation:} Home office spaces are generated based on work-from-home status
\item \textbf{Persona-Specific Asset Placement:} Assets are selected to match the activity behavior of household members.
\end{enumerate}

\subsection{Iterative Bidirectional Influence}

The core innovation of our framework lies in the systematic sharing of semantic information between the two generation modules through an iterative refinement process.

Initially, both the Environment Generation and Human Activity Generation modules receive identical persona information, establishing an \textit{a priori} semantic relationship between anticipated agent activities and environmental object presence based on persona characteristics. 
We establish a system that facilitates iterative bidirectional information flow between the two generation modules. The Environment Generation module produces object inventories, spatial layouts, and affordance maps that constrain and guide activity generation. Conversely, generated activity sequences inform object placement decisions, room utilization patterns, and environmental modifications in subsequent iterations.

\textbf{Convergence Criteria:} The iterative process terminates based on either: (i) reaching a maximum iteration count, or (ii) satisfying user-specified convergence criteria. The user-specified criteria comprise a weighted combination of:

\begin{enumerate}
\item \textbf{Environmental Object Density:} Quantitative measure of object placement density within the environment
\item \textbf{Activity Schedule Granularity:} Density and granularity metrics of the generated activity schedule
\item \textbf{Semantic Similarity Score:} Cosine similarity between environment descriptions and activity descriptions in the SBERT embedding space
\end{enumerate}

\textbf{Algorithm Psuedo code:} The Bidirectional influence controller is described in Algorithm 1. The following symbols are employed throughout the algorithmic specifications:

\begin{itemize}
\item $P$: PersonaData containing demographic and behavioral characteristics
\item $E$: EnvironmentConstraints specifying spatial and functional requirements
\item $Env_i$: Environment schema at iteration $i$
\item $Act_i$: Activity sequences at iteration $i$
\item $N$: Maximum number of iterations for the bidirectional refinement process
\item $\theta$: Convergence threshold for termination criteria
\item $w_j$: Weight parameters for convergence criteria, where $\sum_{j=1}^{4} w_j = 1$
\item $\rho_{env}(i)$: Environmental object density metric at iteration $i$
\item $\gamma_{act}(i)$: Activity schedule granularity metric at iteration $i$
\item $\sigma_{sem}(i)$: Semantic similarity score at iteration $i$
\end{itemize}

The complete algorithmic specification of this iterative bidirectional influence process is presented in Algorithms~1-3.

\begin{algorithm}
\caption{Bidirectional Influence Controller}
\label{alg:bidirectional_controller}
\begin{algorithmic}[1]
\REQUIRE PersonaData $P$, EnvironmentConstraints $E$, MaxIterations $N$, ConvergenceThresholds $T = \{w_1, w_2, w_3, w_4, \theta\}$
\ENSURE EnvironmentSchema $Env$, ActivitySequences $Act$
\STATE $Env_0 \leftarrow EnvGenerator(P, E)$
\STATE $Act_0 \leftarrow ActGenerator(P)$
\STATE $i \leftarrow 0$, $converged \leftarrow false$
\WHILE{$i < N$ AND $\neg converged$}
    \STATE $Env_{i+1} \leftarrow ActToEnvInfluence(Act_i, Env_i, P)$
    \STATE $Act_{i+1} \leftarrow EnvToActInfluence(Env_{i+1}, Act_i, P)$
    \STATE $\rho_{env}(i+1) \leftarrow \frac{|Objects_{i+1}|}{|Rooms_{i+1}|}$
    \STATE $\gamma_{act}(i+1) \leftarrow \frac{\sum_{j} duration_j}{|Activities_{i+1}|}$
    \STATE $\sigma_{sem}(i+1) \leftarrow \frac{embed(Env_{i+1}) \cdot embed(Act_{i+1})}{||embed(Env_{i+1})|| \times ||embed(Act_{i+1})||}$
    \STATE $Score \leftarrow + w_2 \rho_{env}(i+1) + w_3 \gamma_{act}(i+1) + w_4 \sigma_{sem}(i+1)$
    \STATE $converged \leftarrow (Score \geq \theta)$
    \STATE $i \leftarrow i + 1$
\ENDWHILE
\RETURN $Env_i$, $Act_i$
\end{algorithmic}
\end{algorithm}

\begin{table*}[!t]
\centering
\caption{Mutual Information Gain Analysis}
\label{tab:mutual_information}
\begin{tabular}{|c|c|c|c|c|}
\hline
\textbf{Data Points} & \textbf{MI(P,E)+MI(E,B)} & \textbf{MI(P,B)} & \textbf{PCA Components} & \textbf{K-means Clusters} \\
\hline
20 & $0.27 \pm 0.08$ & $0.19 \pm 0.06$ & 14 & 6 \\
50 & $0.48 \pm 0.07$ & $0.30 \pm 0.05$ & 30 & 15 \\
100 & $0.64 \pm 0.06$ & $0.41 \pm 0.04$ & 45 & 17 \\
200 & $0.70 \pm 0.05$ & $0.47 \pm 0.03$ & 51 & 20 \\
\hline
\end{tabular}
\end{table*}

\section{Evaluation}

\subsection{Evaluation Framework}

Our evaluation employs a comprehensive multi-modal embedding approach to validate the effectiveness of bidirectional coupling between persona, environment, and behavioral components. We utilize embeddings of household member personas, environments, and activities: $[persona_{em}, env_{em}, beh_{em}]$ to analyze semantic relationships across modalities using SBERT for textual descriptions and CLIP for visual representations.

We evaluate our framework through three key metrics: (1) \textit{Semantic Alignment Analysis} measuring cosine similarities between component embeddings, (2) \textit{Bidirectional Coupling Validation} using mutual information and iterative improvement measures, and (3) \textit{Persona Responsiveness Analysis} through controlled intervention studies. Additionally, we conduct a comprehensive \textit{Real-World Alignment Validation} experiment comparing our generated data against established datasets using information-theoretic metrics.

\subsection{Statistical Validation}

\textbf{Data Embeddings and Dimensionality Reduction:} We utilize multi-modal embeddings $[persona_{em}, env_{em}, beh_{em}]$ to analyze semantic relationships across components. Persona characteristics are embedded using SBERT, while environments use both SBERT (textual descriptions) and CLIP (visual floorplan representations). Activity sequences and human-robot interactions are embedded to capture behavioral patterns.

To ensure statistical validity, embedding descriptions are standardized by removing household member names, including specific asset descriptions, and using generic room names to minimize noise from irrelevant variations. Prior to analysis, we reduce dimensionality using PCA (n=10 components) or K-means clustering (n=50 clusters):

\begin{eqnarray}
per_{red} &=& \mathrm{PCA}(n=10).fit\_trans(sbert_{per}) \\
env_{red} &=& \mathrm{PCA}(n=10).fit\_trans(sbert_{env})
\end{eqnarray}

This dimensionality reduction ensures consistent feature spaces across baseline and intervention conditions for meaningful statistical comparisons.

\subsubsection{Semantic Alignment Analysis}

We compute pairwise cosine similarities between embedding pairs to measure semantic alignment: $\mathrm{sim}(x, y) = \frac{x \cdot y}{||x|| \times ||y||}$. Tables~\ref{tab:cosine_sbert} and \ref{tab:cosine_clip} present comprehensive results showing strong semantic coherence across all component pairs.

\begin{table}[!t]
\centering
\begin{minipage}{0.48\textwidth}
\centering
\caption{SBERT Cosine Similarity Analysis}
\label{tab:cosine_sbert}
\begin{tabular}{|l|c|}
\hline
\textbf{Embedding Pair} & \textbf{SBERT Similarity} \\
\hline
Persona-Environment & $0.68 \pm 0.09$ \\
Environment-Behavior & $0.72 \pm 0.07$ \\
Persona-Behavior & $0.61 \pm 0.12$ \\
\hline
\end{tabular}
\end{minipage}
\hfill
\begin{minipage}{0.48\textwidth}
\centering
\caption{CLIP Visual-Textual Similarity Analysis}
\label{tab:cosine_clip}
\begin{tabular}{|l|c|}
\hline
\textbf{Comparison} & \textbf{CLIP Similarity} \\
\hline
House Images vs. Household Desc. & $0.74 \pm 0.08$ \\
\hline
\end{tabular}
\end{minipage}
\end{table}

The results demonstrate strong semantic alignment across all component pairs, with Environment-Behavior showing the highest correlation (0.72), indicating effective bidirectional influence. The CLIP visual-textual similarity of 0.74 confirms that generated environments accurately reflect household characteristics.

\subsubsection{Bidirectional Coupling Validation}

We validate bidirectional influence using mutual information to quantify information sharing between components. The framework satisfies the mediation criterion: $MI(persona, env) + MI(env, beh) \textgreater{} MI(persona, beh)$, confirming that environment serves as an effective mediator between persona characteristics and behavioral patterns. Table~\ref{tab:mutual_information} presents detailed mutual information analysis across different data point scales, demonstrating consistent mediation effects with increasing sample sizes.

Table~\ref{tab:iterative_improvement} demonstrates progressive enhancement in both mutual information gain and cosine similarity across iterations, validating the effectiveness of our iterative coupling mechanism. The consistent improvement from iteration 1 (MI: 0.45) to iteration 5 (MI: 0.85) confirms that bidirectional refinement successfully enhances semantic coherence.

\begin{table}[!t]
\centering
\caption{Iterative Improvement Validation Results}
\label{tab:iterative_improvement}
\begin{tabular}{|c|c|c|}
\hline
\textbf{Iter.} & \textbf{MI(P,E)+MI(E,B)} & \textbf{Cosine Sim.} \\
\hline
1 & $0.45 \pm 0.09$ & $0.58 \pm 0.12$ \\
2 & $0.62 \pm 0.08$ & $0.65 \pm 0.10$ \\
3 & $0.74 \pm 0.06$ & $0.71 \pm 0.08$ \\
4 & $0.81 \pm 0.05$ & $0.76 \pm 0.07$ \\
5 & $0.85 \pm 0.04$ & $0.79 \pm 0.06$ \\
\hline
\end{tabular}
\end{table}

\subsubsection{Persona Responsiveness Analysis}

We employ controlled intervention analysis to validate causal relationships between persona characteristics and generated content. We systematically modify three key attributes: age (teenager vs. retiree), organization level (messy vs. organized), and sleep habits (early vs. late), generating N=30 samples for each.

\begin{table*}[!t]
\centering
\caption{Intervention Analysis Statistical Results}
\label{tab:intervention_results}
\begin{tabular}{|l|c|c|c|c|}
\hline
\textbf{Intervention} & \textbf{t-test (p-value)} & \textbf{ANOVA (F-stat)} & \textbf{Cohen's d} & \textbf{Effect Size} \\
\hline
Age: Teenager & $p < 0.001$ & $F = 12.4$ & $d = 0.89$ & Large \\
Age: Retiree & $p < 0.001$ & $F = 15.2$ & $d = 1.12$ & Large \\
Org: Messy & $p = 0.003$ & $F = 8.7$ & $d = 0.64$ & Medium \\
Org: Organized & $p = 0.001$ & $F = 10.1$ & $d = 0.73$ & Medium \\
Sleep: Early & $p = 0.012$ & $F = 6.2$ & $d = 0.51$ & Medium \\
Sleep: Late & $p = 0.008$ & $F = 7.1$ & $d = 0.58$ & Medium \\
\hline
\end{tabular}
\end{table*}

\begin{figure*}[t]
    \centering
    \begin{subfigure}[b]{0.32\textwidth}
        \centering
        \includegraphics[width=\textwidth,height=3cm]{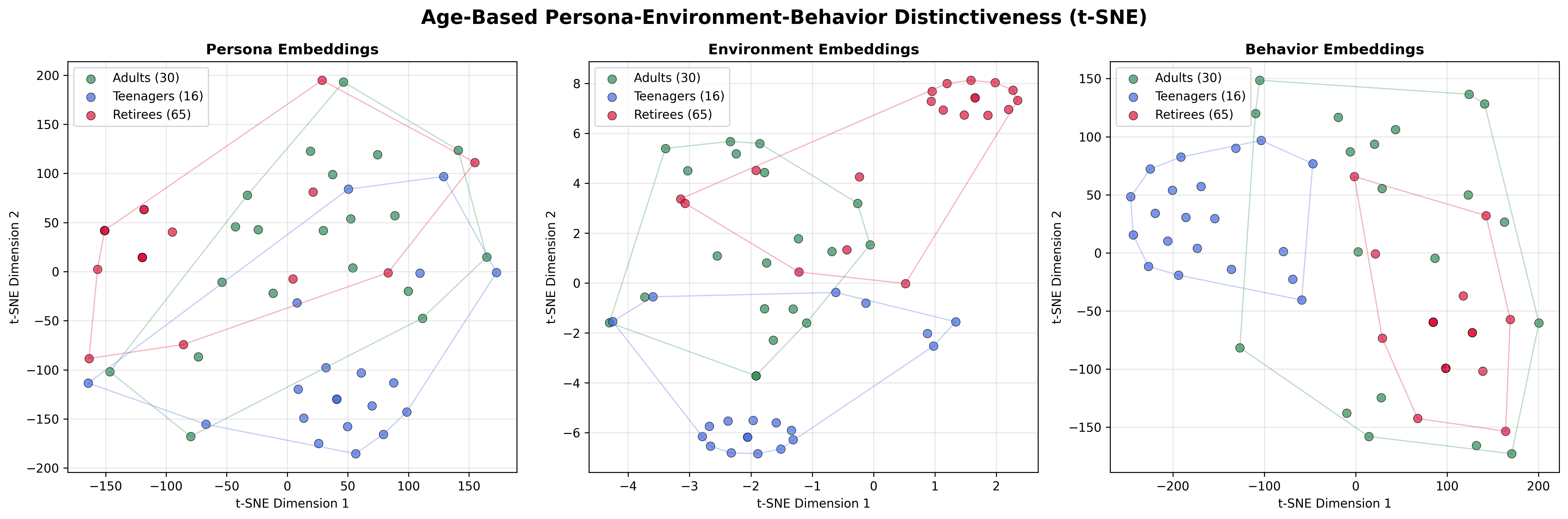}
        \caption{Age Intervention Clusters}
        \label{fig:age_clusters}
    \end{subfigure}
    \hfill
    \begin{subfigure}[b]{0.32\textwidth}
        \centering
        \includegraphics[width=\textwidth,height=3cm]{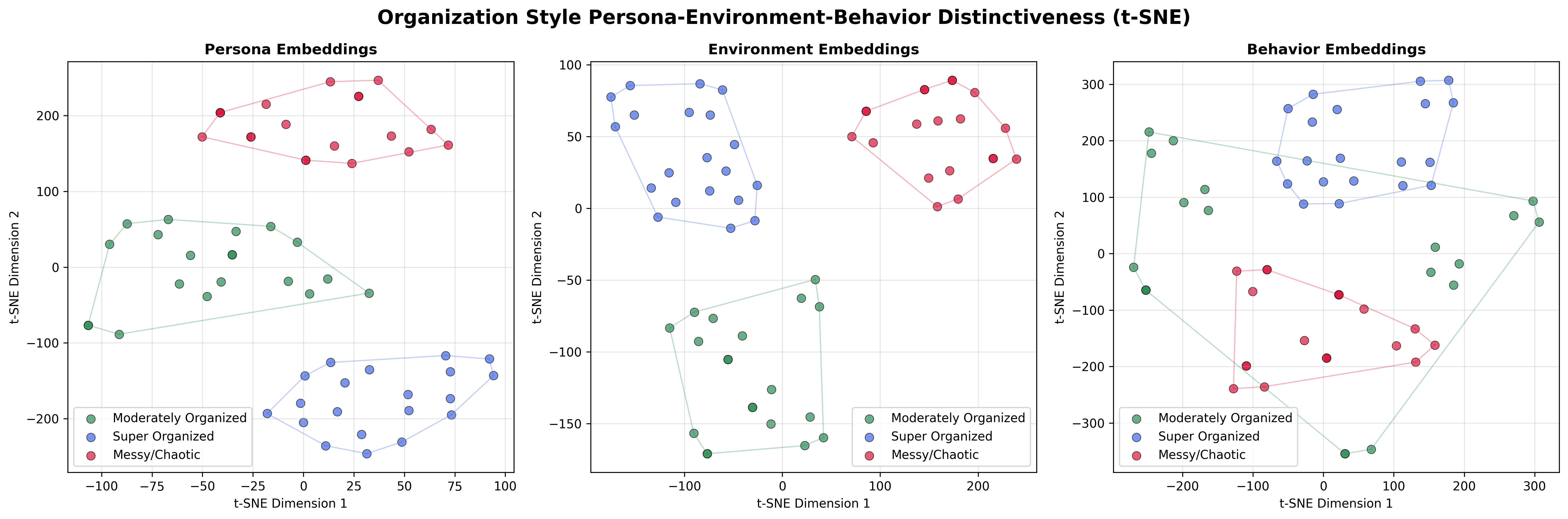}
        \caption{Organization Intervention Clusters}
        \label{fig:org_clusters}
    \end{subfigure}
    \hfill
    \begin{subfigure}[b]{0.32\textwidth}
        \centering
        \includegraphics[width=\textwidth,height=3cm]{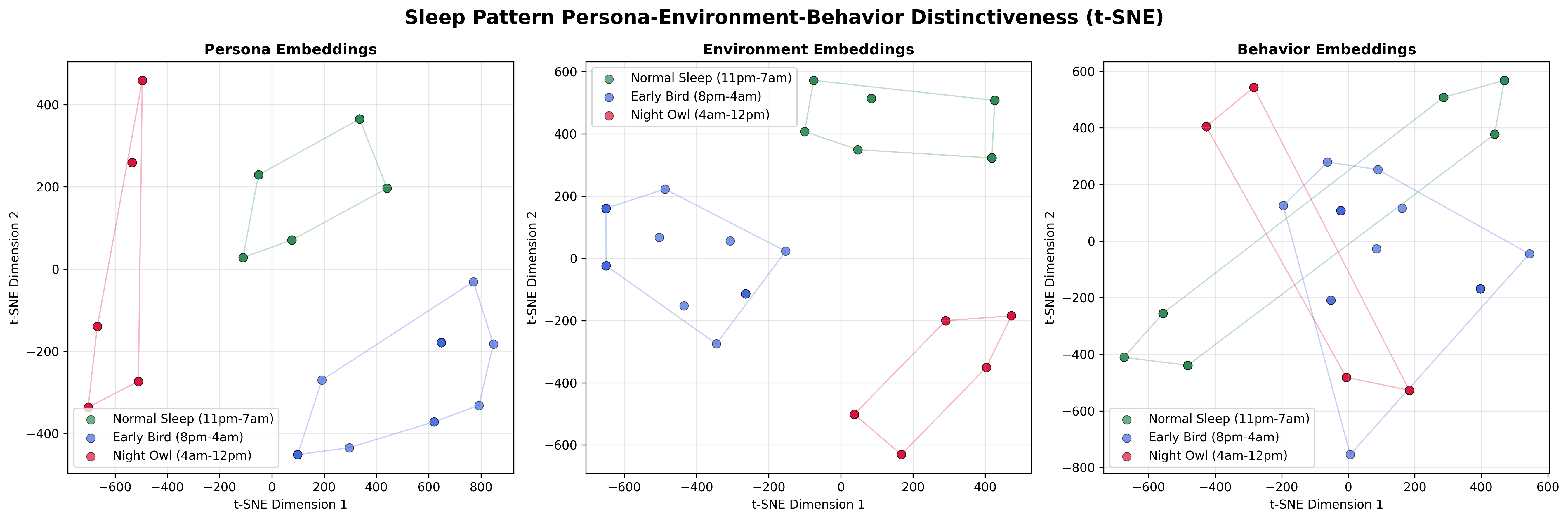}
        \caption{Sleep Pattern Intervention Clusters}
        \label{fig:sleep_clusters}
    \end{subfigure}
    \caption{Intervention Analysis t-SNE Cluster Visualizations: Clear cluster separation validates the framework's ability to generate distinct persona-driven patterns across different intervention conditions.}
    \label{fig:intervention_clusters}
\end{figure*}

All interventions achieve high statistical significance ($p < 0.001$ for age, $p < 0.012$ for others) with substantial effect sizes (Cohen's d = 0.51-1.12), confirming that the bidirectional coupling mechanism successfully translates persona changes into measurable differences in both environmental configurations and behavioral patterns. Figure~\ref{fig:intervention_clusters} visualizes these effects through t-SNE cluster analysis, showing distinct, separable clusters for each intervention type.

\subsection{Real-World Alignment Validation}

We evaluate our framework's ability to generate realistic human behavior patterns by comparing against established datasets: HOMER \cite{patel2022proactiverobotassistancespatiotemporal} (self-reported activity schedules from 21 participants) and Wang et al. \cite{wang2024dynamic} (using 16 of the publicly available synthetically generated personas with 5-day activity sequences).

\textbf{Human Activity Pattern Analysis:}  Our framework processes temporal patterns across real-world reference datasets and synthetic behavioral sequences. We sample temporal activity patterns capturing detailed timestamp information to construct comprehensive behavioral profiles. K-means clustering is applied over activity features (timing, duration, frequency) to reveal distinct behavioral clusters and extract probability distributions for each activity type, representing their temporal likelihood across 24-hour periods. We then represent the activity patterns are encoded as 18-dimensional hourly probability vectors (6 AM-11 PM) that capture the temporal scheduling characteristics of each persona's daily routine.

\textbf{Experimental Setup and Persona Configuration:} For persona description input we use household descriptions for working families with school-age children representing sufficient complexity to validate our framework's ability to capture realistic human-environment interactions while maintaining experimental control.

\textbf{Evaluation Metric:} We employ cosine similarity to quantify alignment between synthetic behavioral data and real-world reference datasets. \textit{Cosine Similarity} measures semantic alignment between synthetic and real-world behavioral patterns. Values range from -1 to 1, with 0.0-0.3 indicating low similarity, 0.3-0.6 indicating moderate similarity, and 0.6-1.0 indicating high similarity. 

\begin{table}[h] 
\caption{Semantic Comparison of Synthetic vs Real Activity Distributions} 
\centering 
\begin{tabular}{|l|c|} 
 \hline 
\textbf{Dataset Comparison} & \textbf{Cosine Similarity} \\ \hline 
Homer-HumanAI & 0.60 \\ 
Wang et al-HumanAI & 0.27 \\
\hline 
\end{tabular} 
\label{tab:metrics} 
\end{table}

\textbf{Results Interpretation:} Our framework shows differential alignment with real-world and synthetic behavioral patterns as shown in Table-\ref{tab:metrics}. The Homer-HumanAI comparison (0.603) demonstrates good semantic alignment, exceeding the moderate similarity threshold ($>$0.5) and validating modeling of self-reported schedules. The Wang et al-HumanAI comparison (0.27) shows limited alignment, indicating significant differences between synthetic character behaviors and real human activity patterns. This suggests our framework is more effective at capturing authentic human behavioral patterns than synthetic persona schedules.

These results validate our framework's capability to generate contextually appropriate and statistically sound human activity patterns that reflect real-world household dynamics while preserving the unique characteristics of different demographic groups.

\subsection{Computational Performance Evaluation}

To assess practical feasibility for large-scale dataset generation, we analyze the number of LLM calls and timing across framework components, as shown in table-\ref{tab:computational_performance}. These metrics represent environment generation and human-robot interaction generation for a three-member household in a three-room environment over a single day, with bidirectional generation evaluated across five iterations.


\begin{table}[H]
\centering
\caption{Computational Performance Analysis}
\label{tab:computational_performance}
\begin{threeparttable}
\begin{tabular}{|l|c|c|}
\hline
\textbf{Module}& \textbf{LLM Calls} & \textbf{Processing Time (s) } \\
\hline
Environment &  10 & 50.00\\
Human-Robot Int.& 7 & 81.04\tnote{*}  \\
Bi-directional Gen.& 5 & 19.00 \\
\hline
\end{tabular}
\begin{tablenotes}
\small
\item[*] Extended processing time due to larger context window incorporating cumulative human interactions and activities throughout the simulated day.
\end{tablenotes}
\end{threeparttable}
\end{table}

\section{Discussion}



\textbf{Indoor Robotics Applications:} While our framework presents a generic approach to synthetic data generation for robots and human-robot interactions, it was developed for the emerging indoor robotics industry. The bidirectional coupling between human behavior and environmental configuration is crucial for robots to operate safely in diverse real-world settings—including cleaning robots that understand household routines, assistive robots anticipating human needs, and smart home systems that adapt to family dynamics. The framework addresses the critical data scarcity challenge facing companies developing robotics products for mass deployment.


\textbf{Practical Deployment vs. Theoretical Innovation:} Unlike academic approaches that prioritize novel algorithmic contributions, this work emphasizes practical utility for training commercial robots requiring robust performance across diverse household environments. Our focus on scalable data generation and sim-to-real validation reflects industry requirements for deployable solutions rather than academic preferences for methodological novelty.


\textbf{Robustness vs. Performance Metrics:} Rather than optimizing controlled experimental metrics, we evaluate practical utility and expressive capability for real-world deployment. Primary limitations include: \textit{Interaction Conflicts} (simultaneous incompatible activities, e.g., loud music during sleep) and \textit{Hallucinated Activities} (impossible scenarios from LLM generation).

\section{Conclusion}
In this paper, we present a novel framework for generating realistic synthetic household data that captures the complex relationships between environment configurations, human behavioral patterns, and human-robot interactions. Our framework's key contributions include a bidirectional coupling mechanism between environment and activity generation, a persona-driven environment generation approach, and methods for maintaining temporal consistency in human activity simulation while enabling scalable variation generation. Statistical validation demonstrated  alignment between our synthetically generated activity patterns and real behavioral patterns, with cosine similarity values showing good performance for persona-distinctive datasets (0.603) while revealing challenges with datasets containing routine universal activities (0.27), providing valuable insights for future framework refinement. The effectiveness of our bidirectional influence mechanism was confirmed through mutual information gains and significant intervention effects validating causal relationships. Qualitative evaluation showed that generated environments accurately reflect persona characteristics, activities are properly grounded in environmental context, and temporal patterns match real-world behavioral data while maintaining semantic coherence in interactions.

\section{Future Directions}
While our framework demonstrates promising results, we acknowledge limitations including computational overhead from iterative refinement, dependency on LLM performance, scaling challenges with complex household scenarios. Additionally, our evaluation reveals performance variation across behavioral datasets. Future work should focus on enhancing real-time adaptation capabilities, expanding further validation across diverse household types and behavioral complexity, improving computational efficiency, and developing standardized benchmark datasets. This research represents a significant step forward in enabling more robust training of embodied AI systems, better testing of household robots, and enhanced understanding of human-environment interactions through scalable synthetic data generation.
\bibliography{aaai2026}

\clearpage

\appendix
\section{Supplementary Materials}

This appendix provides supplementary materials supporting the main paper, including detailed algorithmic specifications, visual examples of framework outputs, and input processing demonstrations.

\subsection{Statistical Evaluation Algorithms}

This section presents the complete algorithmic specifications for the statistical evaluation methods described in Section 4.1, including mutual information mediation analysis and persona intervention analysis procedures.

\begin{algorithm}
\small
\caption{Persona Intervention Analysis}
\label{alg:intervention_analysis}
\begin{algorithmic}[1]
\REQUIRE BaselinePersonas $P_{base}$, InterventionAttributes $\{age, org, sleep\}$, SampleSize $N=30$
\ENSURE InterventionEffects $\Delta_{effects}$, StatisticalSignificance $p_{values}$, EffectSizes $cohen\_d$
\STATE $interventions \leftarrow \{(age, [16, 65]), (org, [messy, organized]), (sleep, [early, late])\}$
\STATE $\Delta_{effects} \leftarrow \{\}$; $p_{values} \leftarrow \{\}$; $cohen\_d \leftarrow \{\}$
\FOR{each $(attr, values) \in interventions$}
    \FOR{each $v_{new} \in values$}
        \STATE $P_{intervention} \leftarrow \textsc{ModifyAttribute}(P_{base}, attr, v_{new})$
        \STATE $E_{baseline} \leftarrow \{\}$; $B_{baseline} \leftarrow \{\}$; $E_{intervention} \leftarrow \{\}$; $B_{intervention} \leftarrow \{\}$
        \FOR{$i = 1$ to $N$}
            \STATE $env_{base}, beh_{base} \leftarrow \textsc{GenerateData}(P_{base})$; $env_{int}, beh_{int} \leftarrow \textsc{GenerateData}(P_{intervention})$
            \STATE $E_{baseline} \leftarrow E_{baseline} \cup \{\textsc{PCA\_reduce}(env_{base})\}$; $B_{baseline} \leftarrow B_{baseline} \cup \{\textsc{PCA\_reduce}(beh_{base})\}$
            \STATE $E_{intervention} \leftarrow E_{intervention} \cup \{\textsc{PCA\_reduce}(env_{int})\}$; $B_{intervention} \leftarrow B_{intervention} \cup \{\textsc{PCA\_reduce}(beh_{int})\}$
        \ENDFOR
        \STATE $p_{env} \leftarrow \textsc{TTest}(E_{baseline}, E_{intervention})$; $p_{beh} \leftarrow \textsc{TTest}(B_{baseline}, B_{intervention})$
        \STATE $F_{stat} \leftarrow \textsc{ANOVA}([E_{baseline}, E_{intervention}, B_{baseline}, B_{intervention}])$
        \STATE $d_{env} \leftarrow \textsc{CohenD}(E_{baseline}, E_{intervention})$; $d_{beh} \leftarrow \textsc{CohenD}(B_{baseline}, B_{intervention})$
        \STATE $\Delta_{effects}[attr][v_{new}] \leftarrow \{p_{env}, p_{beh}\}$; $p_{values}[attr][v_{new}] \leftarrow F_{stat}$; $cohen\_d[attr][v_{new}] \leftarrow \{d_{env}, d_{beh}\}$
    \ENDFOR
\ENDFOR
\RETURN $\Delta_{effects}$, $p_{values}$, $cohen\_d$
\end{algorithmic}
\end{algorithm}

\begin{algorithm}[H]
\small
\caption{Mutual Information Mediation Analysis}
\label{alg:mutual_info_mediation}
\begin{algorithmic}[1]
\REQUIRE PersonaEmbeddings $P_{em}$, EnvironmentEmbeddings $E_{em}$, BehaviorEmbeddings $B_{em}$
\ENSURE MediationScore $M_{score}$, DirectInfluence $D_{score}$
\STATE $P_{reduced} \leftarrow \textsc{PCA}(n=10).\textsc{fit\_transform}(P_{em})$
\STATE $E_{reduced} \leftarrow \textsc{PCA}(n=10).\textsc{fit\_transform}(E_{em})$; $B_{reduced} \leftarrow \textsc{PCA}(n=10).\textsc{fit\_transform}(B_{em})$
\STATE $MI_{PE} \leftarrow \textsc{MutualInfo}(P_{reduced}, E_{reduced})$; $MI_{EB} \leftarrow \textsc{MutualInfo}(E_{reduced}, B_{reduced})$; $MI_{PB} \leftarrow \textsc{MutualInfo}(P_{reduced}, B_{reduced})$
\STATE $M_{score} \leftarrow MI_{PE} + MI_{EB}$; $D_{score} \leftarrow MI_{PB}$
\STATE $mediation\_validated \leftarrow (M_{score} \textgreater{} D_{score})$; $mediation\_strength \leftarrow \frac{M_{score} - D_{score}}{M_{score} + D_{score}}$
\RETURN $M_{score}$, $D_{score}$, $mediation\_validated$, $mediation\_strength$
\end{algorithmic}
\end{algorithm}

\end{document}